%% file: ms.tex
\documentclass{article}
\pdfoutput=1

\usepackage[comma,authoryear]{natbib}
\usepackage{amsmath}
\usepackage{amssymb}
\usepackage{graphicx}
\usepackage{subfig}
\usepackage{url}

\title{Where to put the Image in an Image Caption Generator}
\author{
	Marc Tanti \qquad Albert Gatt\\
	Institute of Linguistics and Language Technology\\
	University of Malta\\
	marc.tanti.06@um.edu.mt \qquad albert.gatt@um.edu.mt
	\and
	Kenneth P. Camilleri\\
	Department of Systems and Control Engineering\\
	University of Malta\\
	kenneth.camilleri@um.edu.mt
}
\date{12-Mar-2018}

\newcommand\etc{{\it etc}}

\DeclareMathOperator{\ReLU}{ReLU}
\DeclareMathOperator{\sig}{sig}
\DeclareMathOperator{\softmax}{softmax}

\begin{document}

\label{firstpage}

\maketitle

\begin{abstract}
	When a recurrent neural network language model is used for caption generation, the image information can be fed to the neural network either by directly incorporating it in the RNN -- conditioning the language model by `injecting' image features -- or in a layer following the RNN -- conditioning the language model by `merging' image features. While both options are attested in the literature, there is as yet no systematic comparison between the two. In this paper we empirically show that it is not especially detrimental to performance whether one architecture is used or another. The merge architecture does have practical advantages, as conditioning by merging allows the RNN's hidden state vector to shrink in size by up to four times. Our results suggest that the visual and linguistic modalities for caption generation need not be jointly encoded by the RNN as that yields large, memory-intensive models with few tangible advantages in performance; rather, the multimodal integration should be delayed to a subsequent stage.
\end{abstract}

\input{tex/introduction}

\input{tex/background}

\input{tex/architectures}

\input{tex/experiments}

\input{tex/results}

\input{tex/conclusion}

\section{Acknowledgements}

The research in this paper is partially funded by the Endeavour Scholarship Scheme (Malta). Scholarships are part-financed by the European Union - European Social Fund (ESF) - Operational Programme II – Cohesion Policy 2014-2020 “Investing in human capital to create more opportunities and promote the well-being of society”.

\bibliographystyle{apalike}
\bibliography{ms}

\label{lastpage}

\end{document}

%% file: tex/introduction.tex
\section{Introduction}
\label{sec:introduction}

Image caption generation\footnote{Throughout this paper we refer to textual descriptions of images as captions, although technically a caption is text that complements an image with extra information that is not available from the image. Specifically, the descriptions we talk about are `concrete' and `conceptual' image descriptions \citep{Hodosh2013}.} is the task of generating a natural language description of the content of an image \citep{Bernardi2016}, also known as a caption. One way to do this is to use a neural language model, typically in the form of a recurrent neural network, or RNN, which is used to generate text (illustrated in Figure~\ref{fig:langmod}). Given a sentence prefix, a neural language model will predict which words are likely to follow. With a small modification, this simple model can be extended into an image caption generator, that is, a language model whose predictions are conditioned on image features. To do this, the neural language model must somehow accept as input not only the sentence prefix, but also the image being captioned. This raises the question: {\em At which stage should image information be introduced into a language model?}

Recent work on image captioning has answered this question in different ways, suggesting different views of the relationship between image and text in the caption generation task. To our knowledge, however, these different models and architectures have not been systematically compared. Yet, the question of where image information should feature in captioning is at the heart of a broader set of questions concerning how language can be grounded in perceptual information, questions which have been addressed by cognitive scientists \citep{Harnad1990} and AI practitioners \citep{Roy2005}. 

As we will show in more detail in Section~\ref{sec:background}, differences in the way caption generation architectures treat image features can be characterised in terms of three distinct sets of design choices:

\begin{figure}[t]
	\centering
	\includegraphics[scale=0.75]{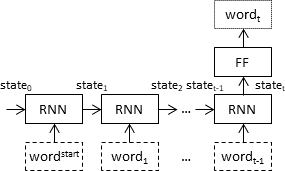}
	\caption{
		\label{fig:langmod}
		How RNN-based neural language models work. Legend: RNN - recurrent neural network; FF - feed forward layer; $\text{word}_i$ - the $i^\text{th}$ generated word in the text; $\text{word}^\text{start}$ - the START token which is an artificial word placed at the beginning of every sentence in order to still have a prefix when predicting the first word (likewise there is an END token to predict the end of a sentence). Note that $\text{state}_1$ represents the prefix `$\text{word}^\text{start}$', $\text{state}_2$ represents the prefix `$\text{word}^\text{start}$ $\text{word}_1$', \etc. After processing a prefix, the RNN passes its final state $\text{state}_t$ to a feedforward layer which then predicts how likely each known word is to be the next word in the prefix.
	}
\end{figure}

\begin{figure}[t]
	\centering
	\subfloat[
	\label{fig:conditioning_inject}
	Conditioning by injecting the image means injecting the image into the same RNN that processes the words.
	]{	\includegraphics[scale=0.75]{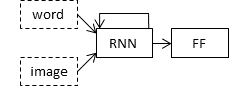}
	}
	\qquad
	\subfloat[
	\label{fig:conditioning_merge}
	Conditioning by merging the image means merging the image with the output of the RNN after processing the words.
	]{
		\includegraphics[scale=0.75]{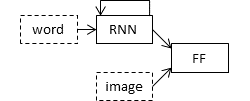}
	}
	\caption{
		\label{fig:conditioning}
		The inject and merge architectures for caption generation. Legend: RNN - recurrent neural network; FF - feed forward layer.
	}
\end{figure}

\paragraph{Conditioning by injecting versus conditioning by merging:} A neural language model can be conditioned by injecting the image (Figure~\ref{fig:conditioning_inject}) or by merging the image (see Figure~\ref{fig:conditioning_merge}). In `inject' architectures, the image vector (usually derived from the activation values of a hidden layer in a convolutional neural network) is injected into the RNN, for example by treating it on a par with a `word' and including it as part of the caption prefix. The RNN is trained to encode the image-language mixture into a single vector in such a way that this vector can be used to predict the next word in the prefix. On the other hand, in the case of `merge' architectures, the image is left out of the RNN subnetwork, such that the RNN handles only the caption prefix, that is, handles only purely linguistic information. After the prefix has been encoded, the image vector is then merged with the prefix vector in a separate `multimodal layer' which comes after the RNN subnetwork. Merging can be done by, for example, concatenating the two vectors together. In this case, the RNN is trained to only encode the prefix and the mixture is handled in a subsequent feedforward layer.

In the terminology adopted in this paper: {\em if an RNN's hidden state vector is somehow influenced by both the image and the words then the image is being injected, otherwise it is being merged}.

\paragraph{Early versus late inclusion of image features:} As the foregoing description suggests, merge architectures tend to incorporate image features somewhat late in the generation process, that is, after processing the whole caption prefix. On the other hand, some inject architectures tend to incorporate image features early in the generation process. Other inject architectures incorporate image features for the whole duration of the generation process. Different architectures can make visual information influence linguistic choices at different stages.

\paragraph{Fixed versus modifiable image features:} For each word predicted, some form of visual information must be available to influence the likelihood of each word. Merge architectures typically use the exact same image representation for every word output. On the other hand, injecting the image features into the RNN allows the internal representation of the image inside the hidden state vector to be changed by the RNN's internal updates after each time step. Different architectures allow for different degrees of modification in the image features for each generated word.
\\

The main contribution of this paper is to present a systematic comparison of the different ways in which the `conditioning' of linguistic choices based on visual information can be carried out, studying their implications for caption generator architectures. Thus, rather than seeking new results that improve on the state of the art, we seek to determine, based on an exhaustive evaluation of inject and merge architectures on a common dataset, where image features are best placed in the caption generation and image retrieval process.\footnote{All the code used in our experiments is available at \url{https://github.com/mtanti/where-image2}.}

From a scientific perspective, such a comparison would be useful for shedding light on the way language can be grounded in vision. Should images and text be intermixed throughout the process, or should they initially be kept separate before being combined in some multimodal layer? Many papers speak of RNNs as `generating' text. Is this the case or are RNNs better viewed as encoders which vectorise a linguistic prefix so that the next feedforward layer can predict the next word, conditioned on an image? Answers to these questions would help inform theories of how caption generation can be performed. The architectures we compare provide different answers to these questions. Hence, it is important to acquire some insights into their relative merits.

From an engineering perspective, insights into the relative performance of different models could provide rules of thumb for selecting an architecture for the task of image captioning, possibly for other tasks as well such as  machine translation. This would make it easier to develop new architectures and new ways to perform caption generation.

The remainder of this paper is structured as follows. We first give an overview of published caption generators based on neural language models, focusing in particular on the architectures used. Section~\ref{sec:architectures} discusses the architectures we compare, followed by a description of the data and experiments in Section~\ref{sec:experiments}. Results are presented and discussed in Section~\ref{sec:results}. We conclude with some general discussion and directions for future work.

%% file: tex/background.tex
\section{Background}
\label{sec:background}

In this section we discuss a number of recent image caption generation models with emphasis on how the image conditions the neural language model, based on the distinction between inject and merge architectures illustrated in Figure~\ref{fig:conditioning}. Before we discuss these models, we first outline four broad sub-categories of architectures that we have identified in the literature.

\subsection{Types of architectures}

\begin{figure}[t]
	\centering
	\subfloat[
	\label{fig:architecture_init}
	Init-inject: The image vector is used as an initial hidden state vector for the RNN.
	]{
		\includegraphics[scale=0.6]{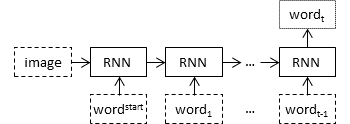}
	}
	\quad
	\subfloat[
	\label{fig:architecture_pre}
	Pre-inject: The image vector is used as a first word in the prefix.
	]{
		\includegraphics[scale=0.6]{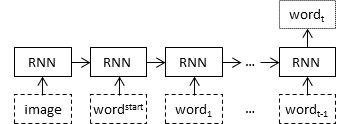}
	}
	
	\subfloat[
	\label{fig:architecture_par}
	Par-inject: The RNN accepts two inputs at once in every time step: a word and an image.
	]{
		\includegraphics[scale=0.6]{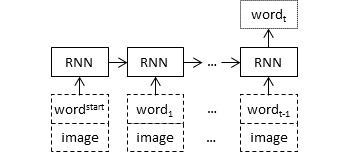}
	}
	\quad
	\subfloat[
	\label{fig:architecture_merge}
	Merge: The image vector is merged with the prefix outside of the RNN.
	]{
		\includegraphics[scale=0.6]{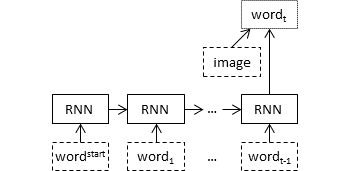}
	}
	\caption{
		\label{fig:architectures}
		Different ways of conditioning a neural language model with an image. The feedforward layer was left out to save space.
	}
\end{figure}

In Section~\ref{sec:introduction}, we made a high-level distinction between architectures that merge linguistic and image features in a multimodal layer, and those that inject image features directly into the caption prefix encoding process. We can in fact distinguish four theoretical possibilities arising from these, as illustrated in Figure~\ref{fig:architectures} and described below.

\begin{itemize}
	\item Init-inject: The RNN's initial hidden state vector is set to be the image vector (or a vector derived from the image vector). It requires the image vector to have the same size as the RNN hidden state vector. This is an early binding architecture and allows the image representation to be modified by the RNN.
	
	\item Pre-inject: The first input to the RNN is the image vector (or a vector derived from the image vector). The word vectors of the caption prefix come later. The image vector is thus treated as a first word in the prefix. It requires the image vector to have the same size as the word vectors. This too is an early binding architecture and allows the image representation to be modified by the RNN.\footnote{In addition to the above, there is an additional, theoretical possibility, which we might refer to as `post-inject'. Post-inject architectures would put the the image vector (or a vector derived from the image vector) at the end of each prefix rather than at the beginning as is done in pre-inject. This would be a late binding architecture which allows minimal modification in the image representation by the RNN. In practice, it would only be possible by structuring the training set as a collection of `sentence prefix - next word' pairs and training the language model using minibatches of individual prefixes rather than full captions at once. No attested work actually adopts this architecture, to our knowledge; hence, we shall not refer to it further in what follows.}
	
	\item Par-inject: The image vector (or a vector derived from the image vector) serves as input to the RNN in parallel with the word vectors of the caption prefix, such that either (a) the RNN takes two separate inputs; or (b) the word vectors are combined with the image vector into a single input before being passed to the RNN. The image vector doesn't need to be exactly the same for each word (such as is the case with attention-based neural models); nor does it need to be included with every word. This is a mixed binding architecture and, whilst allowing some modification in the image representation, it will be harder for the RNN to do so if the same image is fed to the RNN at every time step due to its hidden state vector being refreshed with the original image each time.
	
	\item Merge: The RNN is not exposed to the image vector (or a vector derived from the image vector) at any point. Instead, the image is introduced into the language model after the prefix has been encoded by the RNN in its entirety. This is a late binding architecture and it does not modify the image representation with every time step.
\end{itemize}

\begin{table}
	\centering
	\begin{minipage}{\textwidth}
		\centering
		\begin{tiny}
			\begin{tabular}{lccccp{3cm}}
				\hline\hline
				Source &	Init &	Pre &	Par &	Merge &	Remarks \\
				\hline
				\citep{Chen2014} &	 &	 &	$\checkmark$ &	 &	 \\
				\citep{Chen2015} &	 &	 &	$\checkmark$ &	 &	 \\
				\citep{Devlin2015} &	$\checkmark$ &	 &	 &	 &	 \\
				\citep{Donahue2015} &	 &	 &	$\checkmark$ &	 &	 \\
				\citep{Hendricks2016} &	 &	 &	 &	$\checkmark$ &	 \\
				\citep{Hessel2015} &	 &	 &	$\checkmark$ &	 &	Image is only included with the first word. \\
				\citep{Karpathy2015} &	 &	 &	$\checkmark$ &	 &	Image is only included with the first word. \\
				\citep{Krause2016} &	 &	$\checkmark$ &	 &	 &	Image is passed through a separate RNN at every time step and the hidden state vectors are pre-injected. \\
				\citep{Liu2016}$\dagger$ &	$\checkmark$ &	 &	 &	 &	 \\
				\citep{Liu2016}$\dagger$ &	$\checkmark$ &	 &	$\checkmark$ &	 &	Par-injects image attributes. \\
				\citep{Lu2016} &	 &	 &	$\checkmark$ &	$\checkmark$ &	Attention mechanism which par-injects whole image and merges the attended image.\\
				\citep{Ma2016} &	$\checkmark$ &	 &	 &	 &	Translates image attributes into a caption. \\
				\citep{Mao2014} &	 &	 &	 &	$\checkmark$ &	 \\
				\citep{Mao2015a} &	 &	 &	 &	$\checkmark$ &	 \\
				\citep{Mao2015} &	 &	 &	 &	$\checkmark$ &	 \\
				\citep{Nina2015} &	 &	$\checkmark$ &	 &	 &	 \\
				\citep{Oruganti2016} &	 &	 &	$\checkmark$ &	 &	Image is passed through a separate RNN several times so that a different image hidden state vector is injected at each time step. \\
				\citep{Rennie2016}$\dagger$ &	 &	$\checkmark$ &	 &	 &	 \\
				\citep{Rennie2016}$\dagger$ &	 &	 &	$\checkmark$ &	 &	Attention mechanism which par-injects the attended image into the part of the LSTM that is input gated. \\
				\citep{Vinyals2015} &	 &	$\checkmark$ &	 &	 &	 \\
				\citep{Wang2016} &	$\checkmark$ &	 &	 &	 &	 \\
				\citep{Wu2015} &	 &	$\checkmark$ &	 &	 &	Injects image attributes. \\
				\citep{Xu2015} &	$\checkmark$ &	 &	$\checkmark$ &	$\checkmark$ &	Attention-based mechanism which init-injects the full image while the attended image is par-injected and merged. \\
				\citep{Yao2016}$\dagger$ &	 &	$\checkmark$ &	 &	 &	First two words are the image attributes and the image. \\
				\citep{Yao2016}$\dagger$ &	 &	$\checkmark$ &	$\checkmark$ &	 &	Either pre-inject is made with image attributes and par-inject is made with the image or vice versa. \\
				\citep{You2016} &	 &	$\checkmark$ &	$\checkmark$ &	$\checkmark$ &	 \\
				\citep{Zhou2016} &	 &	$\checkmark$ &	$\checkmark$ &	 &	The image is modified by the last generated word before being par-injected. \\
				\hline\hline
			\end{tabular}
		\end{tiny}
	\end{minipage}
	\caption{
		\label{tbl:litrev_summary}
		Summary of caption generators that use the different conditioning methods. $\dagger$ means that the publication describes multiple systems which use different conditioning methods.
	}
\end{table}

With these distinctions in mind, we next discuss a selection of recent contributions, placing them in the context of this classification. Table~\ref{tbl:litrev_summary} provides a summary of these published architectures.

\paragraph{Init-inject architectures:} Architectures conforming to the init-inject model treat the image vector as the initial hidden state vector of an RNN \citep{Devlin2015,Liu2016}. \citet{Wang2016} combine two RNNs in parallel, both initialized with the same image.

A similar architecture to init-inject is used in traditional deep learning machine translation systems \citep{Sutskever2014} where a source sentence is encoded into a vector and used to condition a language model to generate a sentence in another language. This is the basis for the system described by \citet{Ma2016}, who first extract a sequence of attributes from an image, then translate this sequence into a caption.

It is also used in attention mechanisms in order to provide a vector representing information about the whole image whilst parts of the image that are attended differently during each time step are provided via par-injection. For example \citet{Xu2015} initialize the RNN with the centroid of all image parts before attending to some parts as needed.

\paragraph{Pre-inject architectures:} Pre-inject models treat the image as though it were the first word in the prefix \citep{Vinyals2015,Nina2015,Rennie2016}. Image attributes are sometimes used instead of image vectors \citep{Wu2015,Yao2016}. \citet{Yao2016} also try passing an image as the first two words instead of just one word by using the image vector as the first word and image attributes as a second, or vice versa.

Just like init-inject, pre-inject is also used to provide information about the whole image in attention mechanisms \citep{You2016,Zhou2016}.

\citep{Krause2016} generate paragraph-length captions in two stages. First, an RNN is used to convert the image vector into a sequence of image vectors by incorporating the image at every time step. This sequence of vectors represents sentence topics, each of which is to be converted into a separate sentence by conditioning a language model using pre-inject.

\paragraph{Par-inject architectures:} Par-injection inputs the image features into the RNN jointly with each word in the caption. It is by far the most common architecture used and has the largest variety of implementation. For example \citet{Donahue2015} do this with two RNNs in series and find that it is better to inject the image in the second RNN than the first. \citet{Yao2016} par-inject the image whilst pre-injecting image attributes (or vice versa); and \citet{Liu2016} par-inject attributes from the image whilst init-injecting the image vector. Other, less common instantiations include par-injecting the image, but only with the first word (this is not pre-inject as the image is not injected on a separate time step) \citep{Karpathy2015,Hessel2015}; and passing the words through a separate RNN, such that the resulting hidden state vectors are what is combined with the image vector \citep{Oruganti2016}.

Many times this architecture is used in order to pass a different representation of the same image with every word so that visual information changes for different parts of the sentence being generated. For example \citet{Zhou2016} perform element-wise multiplication of the image vector with the last generated word's embedding vector in order to attend to different parts of the image vector. \citet{Oruganti2016} pass the image through its own RNN for as many times as there are words in order to use a different image vector for every word. \citet{Chen2014,Chen2015} use a simple RNN to try to predict what the image vector looks like given a prefix. This predicted image is then used as a second image representation which is par-injected together with the actual image vector.

More commonly, modified image representations come from attention mechanisms \citep{You2016,Xu2015,Rennie2016}. \citet{Rennie2016} inject the image not as an input to the RNN but use a modified long short term memory network \citep{Hochreiter1997}, or LSTM, which allows them to inject the attended image directly inside the input gated expression (the part of the LSTM which is multiplied by the input gate).

Like init-inject and pre-inject, par-inject is sometimes used to provide information about the whole image in attention mechanisms whilst the attended image regions are merged \citep{Lu2016}.

\paragraph{Merge architectures:} Rather than combining image features together with linguistic features from within the RNN, merge architectures delay their combination until after the caption prefix has been vectorised \citep{Mao2014,Mao2015,Mao2015a}. \citet{Hendricks2016} use a merge architecture in order to keep the image out of the RNN and thus be able to train the part of the neural network that handles images and the part that handles language separately, using images and sentences from separate training sets.

Some work on attention mechanisms also uses merge architectures with attention mechanisms by merging a different image representation at every time step. \citet{You2016} and \citet{Xu2015} merge as well as par-inject the attended visual regions, whilst \citet{Lu2016} only merge the regions whilst par-injecting a fixed image representation.

Though they do not use an RNN and hence are not focussed on in this review, caption generators that use log-bilinear models \citep{Mnih2007} usually merge the image with the prefix representation \citep{Kiros2014,Kiros2014a,Song2016}.

\subsection{Summary and outlook}

While the literature on caption generation now provides a rich range of models and comparative evaluations, there is as yet very little explicit systematic comparison between the performance of the architectures surveyed above, each of which represents a different way of conditioning the prediction of language sequences on visual information. Work that has tested both par-inject and pre-inject, such as by \citet{Vinyals2015}, reports that pre-inject works better. The work of \citet{Mao2015} compares inject and merge architectures and concludes that merge is better than inject. However Mao {\em et al.}'s comparison between architectures is a relatively tangential part of their overall evaluation, and is based only on the BLEU metric \citep{Papineni2002}.

Answering the question of which architecture is best is difficult because different architectures perform differently on different evaluation measures, as shown for example by \citet{Wang2016}, who compared architectures with simple RNNs and LSTMs. Although the state of the art systems in caption generation all use inject-type architectures, it is also the case that they are more complex systems than the published merge architectures and so it is not fair to conclude that inject is better than merge based on a survey of the literature alone.

In what follows, we present a systematic comparison between all the different architectures discussed above. We perform these evaluations using a common dataset and a variety of quality metrics, covering (a) the quality of the generated captions; (b) the linguistic diversity of the generated captions; and (c) the networks' capabilities to determine the most relevant image given a caption.

%% file: tex/architectures.tex
\section{Architectures}
\label{sec:architectures}

\begin{figure}[t]
	\centering
	\includegraphics[scale=0.75]{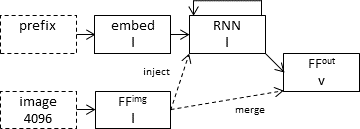}
	\caption{
		\label{fig:implementation}
		An illustration of the main architecture schema that is instantiated in the four different architectures tested in this paper. Legend: `FF' - fully connected feed forward layer with bias; `$\text{FF}^\text{img}$' - layer projecting the image vector (may or may not have an activation function); `$\text{FF}^\text{out}$' - layer projecting into the softmax output; `$l$' - the layer size (which is the same for three different layers); `$v$' - the vocabulary size (which is different for different datasets). Only one of the dashed arrows is used depending on whether the architecture is one of merge or inject.
	}
\end{figure}

In this section we go over the different architectures that are evaluated in this paper. A diagram illustrating the main architecture schema, which is the basis of every tested architecture in this work, is shown in Figure~\ref{fig:implementation}. The schema is based on the architecture described in \citet{Vinyals2015}, without the ensemble. This architecture was chosen for its simplicity whilst still being the best performing system in the 2015 MSCOCO image captioning challenge.\footnote{See: \url{http://mscoco.org/dataset/#captions-leaderboard}}

\paragraph{Word embeddings:} Word embeddings, that is, the vectors that represent known words prior to being fed to the RNN, consist of vectors that have been randomly initialised. No precompiled vector embeddings such as word2vec \citep{Mikolov2013} were used. Instead, the embeddings are trained as part of the neural network in order to learn the best representations of words for the task.

\paragraph{Recurrent neural network:} The purpose of the RNN is to take a prefix of embedded words (with image vector in inject architectures) and produce a single vector that represents the sequence. A gated recurrent unit \citep{Chung2014}, or GRU, was used in our experiments for the simple reason that it is a powerful RNN that only has one hidden state vector. By contrast, an LSTM has two state vectors (hidden and cell states). This would make architecture comparisons more complex, as the presence of two state vectors raise the possibility of multiple versions of the init-inject architecture. By using an RNN with a single hidden state vector there is only one way to implement init-inject.

\paragraph{Image:} Prior to training, all images were vectorised using the activation values of the penultimate layer of the VGG OxfordNet 19-layer convolutional neural network \citep{Simonyan2014}, which is trained to perform object recognition and returns a 4096-element vector. The convolutional neural network is not influenced by the caption generation training. During training, a feed forward layer of the neural network compresses this vector into a smaller vector.

\paragraph{Output:} Once the image and the caption prefix have been vectorised and mixed into a single vector, the next step is to use them to predict the next word in the caption. This is done by passing the mixed vector through a feed-forward layer with a softmax activation function that outputs the probability of each possible next word in the vocabulary. Based on this distribution, the next word that comes after the prefix is selected.

The four architectures discussed in the previous section are evaluated in our experiments as follows:

\begin{itemize}
	\item init-inject: The image vector is treated as an initial hidden state vector for the RNN. After initialising the RNN, the vectors in the caption prefix are then fed to the RNN as usual.
	
	\item pre-inject: The image vector is used as the first `word' in the caption prefix. This makes the image vector the first input that the RNN will see.
	
	\item par-inject: The image vector is concatenated to every word vector in the caption prefix in order to make the RNN take a mixed word-image vector. Every word would have the exact same image vector concatenated to it.
	
	\item merge: The image vector and caption prefix vector are concatenated into a single vector before being fed to the output layer.
\end{itemize}

We now discuss the architecture in a more formal notation. As a matter of notation, we treat vectors as horizontal.

The GRU model is defined as follows:
\begin{align}
r_t &= \sig(x_t W_{xr} + s_{t-1} W_{sr} + b_r) \\
u_t &= \sig(x_t W_{xu} + s_{t-1} W_{su} + b_u) \\
c_t &= \tanh(x_t W_{xc} + (r \odot s_{t-1}) W_{sc} + b_c) \\
s_t &= u_t \odot s_{t-1} + (1 - u_t) \odot c_t
\end{align}
where $x_t$ is the $t^\text{th}$ input, $s_t$ is the hidden state vector after $t$ inputs, $r_t$ is the reset gate after $t$ inputs, $u_t$ is the update gate after $t$ inputs, $W_{\alpha \beta}$ is the weight matrix between $\alpha$ and $\beta$, $b_\alpha$ is the bias vector for $\alpha$, and $\odot$ is the elementwise vector multiplication operator. In the above, `sig' refers to the sigmoid function which is defined as:

\begin{align}
\sig(x) &= \frac{1}{1 + e^{-x}}
\end{align}

The feedforward layers used for the image and output are defined as
\begin{align}
z &= x W + b
\end{align}
where $z$ is the net vector, $x$ is the input vector, $W$ is the weight matrix, and $b$ is the bias vector.

The net vector can then be passed through an activation function, such as the softmax function, which is defined as
\begin{align}
\softmax(z)_i &= \frac{e^{z_i}}{\sum_j e^{z_j}}
\end{align}
where $\softmax(z)_i$ refers to the $i^\text{th}$ element of the new vector.

Another activation function is the rectified linear unit function, or ReLU, which is defined as
\begin{align}
\ReLU(z)_i &= \max(z_i, 0)
\end{align}
where $\ReLU(z)_i$ refers to the $i^\text{th}$ element of the new vector.

%% file: tex/experiments.tex
\section{Experiments}
\label{sec:experiments}

This section describes the experiments conducted in order to compare the performance of the different architectures described in the previous section. Tensorflow\footnote{See: \url{https://www.tensorflow.org/}} v1.2 was used to implement the neural networks.

\subsection{Datasets}
\label{sec:dataset}

The datasets used for all experiments were the version of Flickr8K \citep{Hodosh2013}, Flickr30K \citep{Young2014}, and MSCOCO \citep{Lin2014} distributed by \citet{Karpathy2015}.\footnote{See: \url{http://cs.stanford.edu/people/karpathy/deepimagesent/}} All three datasets consist of images taken from Flickr combined with between five and seven manually written captions per image. The provided datasets are split into a training, validation, and test set using the following number of images respectively: Flickr8K - 6000, 1000, 1000; Flickr30K - 29000, 1014, 1000; MSCOCO - 82783, 5000, 5000. The images are already vectorised into 4096-element vectors via the activations of layer `fc7' (the penultimate layer) of the VGG OxfordNet 19-layer convolutional neural network \citep{Simonyan2014}, which was trained for object recognition on the ImageNet dataset \citep{Deng2009}.

The known vocabulary consists of all the words in the captions of the training set that occur at least 5 times. This amounts to 2539 tokens for Flickr8K, 7415 tokens for Flickr30K, and 8792 tokens for MSCOCO. These words are used both as inputs, which are embedded and fed to the RNN, and as outputs, which are assigned probabilities by the softmax function. Any other word which is not part of the vocabulary is replaced with an UNKNOWN token.

\subsection{Hyperparameter tuning}
\label{sec:hyperparameter_tuning}

For the results to be reliable, it is important to find the best (within practical limits) hyperparameters for each architecture so that we can judge the performance of the architectures when they are optimally tuned, rather than using one-size-fits-all hyperparameter settings which might cause some architectures to under-perform. For this reason we used a multi-step process of hyperparameter tuning, which is described below. We optimized the hyperparameters in order to maximize caption quality on the Flickr8K validation set, using beam search as a generation method and CIDEr as the objective function. The optimal hyperparameters were then fixed across all datasets. \citet{Mao2015} also used Flickr8K for hyperparameter tuning and CIDEr was shown by \citet{Rennie2016} to be a useful metric to optimise on, yielding an improvement on other quality metrics when used as the objective function.

The following hyperparameters were fixed across all architectures:

\begin{itemize}
	\item Parameter optimization is performed using the Adam algorithm \citep{P.Kingma2014} with its hyperparameters kept as suggested in the original paper: $\alpha = 0.001$, $\beta_1 = 0.9$, $\beta_2 = 0.999$, and $\epsilon = 10^{-8}$.
	\item The loss function is the mean of the cross-entropy of each word in each caption in a minibatch. The crossentropy of the $t^\text{th}$ word in a caption is defined as follows:
	\begin{align}
	\text{crossentropy}(P, I, C_{0 \dots {t-1}}, C_{t}) = -\ln \left( { P(C_{t} | C_{0 \dots {t-1}}, I) } \right)
	\end{align}
	where $P$ is the trained neural network that gives the probability of a particular word being the next word in a caption prefix, $C$ is a caption with $|C|$ words, and $I$ is an image described by caption $C$. Note that $C_t$ is the $t^\text{th}$ word in $C$ and $C_{0 \dots {t-1}}$ are the first $t-1$ words in $C$ plus the START token.
	\item An early stopping criterion is used, such that the geometric mean of the language model perplexity on the validation set is measured and as soon as one epoch results in a worse perplexity than the previous epoch, the training stops. A maximum number of epochs are still used to prevent training from going on for too long (more on this later).
	\item During caption generation, the caption must be between 5 and 50 words long. Beam search will not end a sentence before there are at least 5 words in it and will abruptly stop using a partial sentence that is 50 words long.
	\item All biases are initialized to zeros.
\end{itemize}

\noindent The following are hyperparameters that were tuned (the ranges of values were minimized in order to keep the search space tractable):
\begin{itemize}
	\item The weights initialization procedure (normal distribution or xavier \citep{Glorot2010} with normal distribution).
	\item The weights initialization range ($-0.1$ to $0.1$ or $-0.01$ to $0.01$).
	\item The size of the layers for embedding, image projection ($\text{FF}^\text{img}$ in Figure~\ref{fig:implementation}), and RNN hidden state vector (64, 128, 256, or 512), all three of which are constrained to be equal\footnote{Note that if we allowed each layer to change freely from the other layers, init-inject would still require that the image size and RNN size be equal and pre-inject would still require that the image size and the embedding size be equal, whilst par-inject and merge would have no such size restrictions. This would make the former two architectures have significantly less hyperparameter combinations to explore which would likely result in an unfair advantage after hyperparameter tuning.}.
	\item Whether to normalize the image vector before passing it to the neural network.
	\item Whether to use ReLU after the image projection ($\text{FF}^\text{img}$ in Figure~\ref{fig:implementation}) or to leave it linear.
	\item Whether to use an all-zeros vector as an initial RNN hidden state vector or to use a learnable vector (not applicable to init-inject since its initial hidden state vector is the image projection).
	\item Whether to use L2 weights regularization with a weighting constant of $10^{-8}$.
	\item Whether to apply dropout regularisation at different points in the architecture (in Figure~\ref{fig:implementation}: after `image', after `$\text{FF}^\text{img}$', after `embed', and/or after `RNN'). Each application of dropout (if any) has a dropout rate of 0.5.
	\item The minibatch size (32, 64, or 128).
\end{itemize}

\noindent The following steps were followed in order to tune these hyperparameters, which were evaluated by training a neural network for a maximum of 10 epochs, generating captions with a beam width of 2, and evaluating the captions using CIDEr:

\begin{itemize}
	\item[1.] Randomly generate 100 unique hyperparameter combinations and record their performance.
	\item[2.] Use Baysian optimization via the library GPyOpt\footnote{See: \url{http://sheffieldml.github.io/GPyOpt/}} for 100 iterations and record each generated candidate combination's performance. Use the combinations from step 1 to initialize the search.
	\item[3.] Use trees of Parzan estimators via the library hyperopt\footnote{See: \url{https://jaberg.github.io/hyperopt/}} for 100 iterations and record each generated candidate combination's performance.
	\item[4.] Take the best combination found in all of the previous steps and fine-tune it using greedy hill climbing and record each modified combination. This is to check if changing any one hyperparameter will improve the performance.
\end{itemize}

\noindent The previous steps do not have very reliable CIDEr scores associated with them as their score was produced using just one training and generation run and so might coincidentally be an unusual score (far from the mean score if we trained the same neural network several times). Ideally we would have tested each hyperparameter combination three times and taken the mean of the resulting CIDEr scores. Ideally we would have also tried different values for maximum number of epochs and beam width. This, however, would have been extremely time consuming. Thus, we only apply the procedure to a subset of the best performing combinations from the previous steps. We ensure that the subset is diverse by only choosing combinations that are dissimilar from each other, as follows:

\begin{itemize}
	\item[5.] Take all duplicate combinations generated in all of the previous steps and replace them with a single combination with their average CIDEr score. Take the top 10 scoring combinations.
	\item[6.] Out of the selected 10 combinations take the three combinations that are most different from each other in terms of Hamming distance. Ensure that one of these three combinations is the best combination found in the previous step.
	\item[7.] Take the three combinations selected and try different maximum epochs (10 and 100) and beam widths (1, 2, 3, 4, 5, and 6) on them. Each evaluation is measured using the average CIDEr score of three independent training and generation runs.
	\item[8.] Return the best combination found in the previous step.
\end{itemize}

In Section~\ref{sec:results:hyperparameters} we will discuss the optimal hyperparameters found.

\subsection{Evaluation metrics}

To evaluate the different architectures, the test set captions (which are shared among all architectures) are used to measure the architectures' quality using metrics that fall into three classes, described below.

\paragraph{Generation metrics:} These metrics quantify the quality of the generated captions by measuring the degree of overlap between generated captions and those in the test set. We use the MSCOCO evaluation code\footnote{See: \url{https://github.com/tylin/coco-caption}} which measures the standard evaluation metrics BLEU-(1,2,3,4) \citep{Papineni2002}, ROUGE-L \citep{Lin2004}, METEOR \citep{Banerjee2005}, and CIDEr \citep{Vedantam2015}.

\paragraph{Diversity metrics:} Apart from measuring the caption similarity to the ground truth we also measure the diversity of the vocabulary used in the generated captions. This is intended to shed light on the extent to which the captions produced by models are `stereotyped', that is, the extent to which a model re-uses (sub-)strings from case to case, irrespective of the input image.

As a limiting case, consider a caption generator which always outputs the same caption. Such a generator would have the lowest possible diversity score. In order to quantify this we measure the percentage of known vocabulary words used in all generated captions and the entropy of the unigram and bigram frequencies in all the generated captions together, which is calculated as:
\begin{align}
\text{entropy}(F) &= -\sum_{i=1}^{|F|} { P_i(F) \log_2{P_i(F)} } \\
P_i(F) &= \frac{F_i}{\sum_{j=1}^{|F|} {F_j}}
\end{align}
where $F$ is the frequency distribution over generated unigrams or bigrams with $|F|$ different types of unigrams or bigrams and $P_i$ is the maximum likelihood estimate probability of encountering unigram or bigram $i$. Note that $F_i$ is the frequency of the unigram or bigram $i$.

Entropy gives a measure of how uniform the frequency distributions are (with higher entropy for more uniform distributions). The more uniform, the more likely that each unigram or bigram was used in equal proportion, rather than using the same few words for the majority of the time, hence the greater the variety of words used.

Finally we also measure the percentage of generated captions that already exist in the training set, as an estimate of the extent to which a model evinces `parroting', or wholesale caption reuse from the training set.

For these diversity metrics, we obtain a ceiling estimate by computing the same measures on the test set captions themselves. We take the first caption out of the group of human-written captions available for each image in the test set and apply these diversity metrics on them.

\paragraph{Retrieval metrics:} Retrieval metrics are metrics that quantify how well the architectures perform when retrieving the correct image out of all the test set images in the test set given a corresponding caption. A conditioned language model can be used for retrieval by measuring the degree of relevance each image has to the given caption. Relevance is measured as the probability of the whole caption given the image (by multiplying together each word's probability). Different images will give different probabilities for the same caption. The more probable the caption is, the more relevant the image.

We use the standard $R$@$n$ recall measures \citep{Hodosh2013}, and report recall at 1, 5, and 10. Recall at $n$ is the percentage of captions whose correct image is among the top $n$ most relevant images.

Since this process takes time proportional to the number of captions multiplied by the number of images, the pool of possible captions to consider during retrieval excluded all captions except the first out of the group of captions available for each image in order to reduce the evaluation time. For MSCOCO we only used the first 1000 test set images out of 5000 for the same reason, similar to Flickr8K and Flickr30K which only have 1000 images.

We also included the language model perplexity. The perplexity of a sentence/image pair is calculated as:
\begin{align}
\text{perplexity}(P, C, I) &= 2^{H(P, C, I)} \\
H(P, C, I) &= -\frac{1}{|C|} \sum_{n=0}^{|C|} { \log_2 \left( P(C_{t} | C_{0 \ldots {t-1}}, I) \right)}
\end{align}
where $P$ is the trained neural network that gives the probability of a particular word being the next word in a caption prefix, $C$ is a caption with $|C|$ words, $I$ is an image described by caption $C$, and $H$ is the entropy function. Note that $C_t$ is the $t^\text{th}$ word in $C$ and $C_{0 \dots {t-1}}$ are the first $t-1$ words in $C$ plus the START token.

In order to aggregate the caption perplexity of the entire test set of captions into a single number, we report the geometric mean of all the caption's scores.

%% file: tex/results.tex
\section{Results and discussion}
\label{sec:results}

Three runs of each experiment, on each of the three datasets, were performed. For the various evaluation measures, we report the mean together with the standard deviation (reported in parentheses) over the three runs. For each run, the initial model weights, minibatch selections, and dropout selections are different since these are randomly determined. Everything else is identical across runs.

\subsection{Optimal hyperparameters}
\label{sec:results:hyperparameters}

We start by discussing the optimal hyperparameters found for each architecture which are listed in Table~\ref{tbl:hyperparameters}. 

\begin{table}[!h]
	\centering
	\caption{
		\label{tbl:hyperparameters}
		The optimal hyperparameters found for each architecture, tuned on Flickr8K with CIDEr as objective function. Each row is explained in Section~\ref{sec:hyperparameter_tuning}.
	}
	\begin{minipage}{\textwidth}
		\centering
		\begin{small}
			\begin{tabular}{lcccc}
				\hline\hline
				&	init-inject &	pre-inject &	par-inject &	merge \\
				\hline
				init. method &	xavier &	normal &	normal &	normal \\
				init. weight range &	$-0.01$ -- $0.01$ &	$-0.1$ -- $0.1$ &	$-0.1$ -- $0.1$ &	$-0.1$ -- $0.1$ \\
				layer size &	512 &	512 &	256 &	128 \\
				normalize image &	yes &	yes &	yes &	yes \\
				image activation &	none &	none &	none &	none \\
				init. RNN hidden state &	N/A &	zero &	learnable &	learnable \\
				regularize weights &	no &	no &	yes &	no \\
				image dropout &	no &	no &	yes &	no \\
				image proj. dropout &	no &	no &	no &	no \\
				embedding dropout &	yes &	yes &	yes &	no \\
				RNN dropout &	yes &	yes &	yes &	yes \\
				minibatch size &	128 &	32 &	64 &	128 \\
				max. epochs &	100 &	100 &	100 &	100 \\
				beam width &	3 &	3 &	5 &	3 \\
				\hline\hline
			\end{tabular}
		\end{small}
	\end{minipage}
\end{table}

It is interesting to note that, in every architecture's optimal hyperparameters, the RNN output needs to be regularized with dropout, the image vector should not have a non-linear activation function or be regularized with dropout, and the image input vector must be normalized before being fed to the neural network. Par-inject seems to need the most help in terms of regularization and even in terms of beam width, whilst the small size of merge means that it needs the least amount of regularization.

The most interesting observation is that the merge architecture is much `leaner' overall. In terms of RNN size, it needs half of what par-inject needs, and only a quarter of what init-inject and pre-inject require for optimal performance. This makes sense, since merge only needs the RNN for storing linguistic information, whilst the other architectures need to additionally store visual information from the image. Using a larger RNN with the merge architecture would likely lead to overfitting. 

The implication is that init-inject and pre-inject are much more memory-hungry architectures that require large RNN hidden state vectors in order to function well, whilst merge is more efficient. In fact, the number of parameters for merge is between 3 and 4 times smaller than the number of parameters for init-inject and pre-inject. Merge is also about 2 or 3 times faster to train. 

Of the inject architectures, par-inject has the smallest optimal RNN size. This is probably due to the fact that, in this model, the image is present at all time steps, thereby necessitating less memory to be allocated to `remember' visual information together with linguistic information, compared to early-binding architectures. It's interesting to note that the par-inject RNN size is equal to the size of the concatenated image and RNN hidden state vector in the merge architecture.

\subsection{Quality of generated captions}

Table~\ref{tbl:results_gen1} and Table~\ref{tbl:results_gen2} display the metrics that measure the quality of generated captions, calculated using the MSCOCO evaluation toolkit and averaged over the three experimental runs.

\begin{table}[t]
	\centering
	\caption{
		\label{tbl:results_gen1}Results of caption quality metrics (CIDEr, METEOR, and ROUGE-L).
	}
	\subfloat[
		\label{tbl:results_gen_flickr8k_1}Results for Flickr8K.
	]{
		\begin{minipage}{\textwidth}
			\centering
			\begin{small}
				\begin{tabular}{lccc}
					\hline\hline
					&	CIDEr &	METEOR &	ROUGE-L \\
					init-inject &	\bf 0.481 (0.010) &	\bf 0.194 (0.000) &	0.445 (0.002) \\
					par-inject &	0.475 (0.004) &	0.193 (0.002) &	\bf 0.448 (0.003) \\
					pre-inject &	0.469 (0.009) &	0.191 (0.001) &	0.444 (0.003) \\
					merge &	0.469 (0.015) &	0.193 (0.002) &	0.443 (0.003) \\
					\hline\hline
				\end{tabular}
			\end{small}
		\end{minipage}
	}
	
	\subfloat[
		\label{tbl:results_gen_flickr30k_1}Results for Flickr30K.
	]{
		\begin{minipage}{\textwidth}
			\centering
			\begin{small}
				\begin{tabular}{lccc}
					\hline\hline
					&	CIDEr &	METEOR &	ROUGE-L \\
					merge &	\bf 0.385 (0.006) &	0.174 (0.000) &	0.423 (0.001) \\
					init-inject &	0.383 (0.005) &	\bf 0.177 (0.002) &	\bf 0.425 (0.003) \\
					pre-inject &	0.380 (0.006) &	0.174 (0.001) &	0.420 (0.002) \\
					par-inject &	0.361 (0.004) &	0.170 (0.002) &	0.418 (0.001) \\
					\hline\hline
				\end{tabular}
			\end{small}
		\end{minipage}
	}
	
	\subfloat[
		\label{tbl:results_gen_mscoco_1}Results for MSCOCO.
	]{
		\begin{minipage}{\textwidth}
			\centering
			\begin{small}
				\begin{tabular}{lccc}
					\hline\hline
					&	CIDEr &	METEOR &	ROUGE-L \\
					init-inject &	\bf 0.818 (0.005) &	\bf 0.226 (0.002) &	\bf 0.499 (0.003) \\
					pre-inject &	0.807 (0.007) &	0.224 (0.000) &	0.498 (0.002) \\
					merge &	0.791 (0.010) &	0.222 (0.001) &	0.494 (0.002) \\
					par-inject &	0.774 (0.003) &	0.219 (0.001) &	0.493 (0.001) \\
					\hline\hline
				\end{tabular}
			\end{small}
		\end{minipage}
	}
\end{table}

\begin{table}[t]
	\centering
	\caption{
		\label{tbl:results_gen2}Results of caption quality metrics (BLEU-1, BLEU-2, BLEU-3, and BLEU-4).
	}
	\subfloat[
		\label{tbl:results_gen_flickr8k_2}Results for Flickr8K.
	]{
		\begin{minipage}{\textwidth}
			\centering
			\begin{small}
				\begin{tabular}{lcccc}
					\hline\hline
					&	BLEU-4 &	BLEU-3 &	BLEU-2 &	BLEU-1 \\
					par-inject &	\bf 0.191 (0.003) &	\bf 0.287 (0.003) &	\bf 0.424 (0.002) &	\bf 0.611 (0.001) \\
					init-inject &	0.191 (0.004) &	0.285 (0.005) &	0.424 (0.005) &	0.611 (0.002) \\
					pre-inject &	0.190 (0.003) &	0.285 (0.004) &	0.421 (0.005) &	0.609 (0.007) \\
					merge &	0.178 (0.004) &	0.273 (0.005) &	0.413 (0.006) &	0.600 (0.007) \\
					\hline\hline
				\end{tabular}
			\end{small}
		\end{minipage}
	}
	
	\subfloat[
		\label{tbl:results_gen_flickr30k_2}Results for Flickr30K.
	]{
		\begin{minipage}{\textwidth}
			\centering
			\begin{small}
				\begin{tabular}{lcccc}
					\hline\hline
					&	BLEU-4 &	BLEU-3 &	BLEU-2 &	BLEU-1 \\
					pre-inject &	\bf 0.192 (0.002) &	\bf 0.284 (0.001) &	\bf 0.419 (0.003) &	0.613 (0.004) \\
					init-inject &	0.191 (0.002) &	0.283 (0.002) &	0.419 (0.002) &	0.613 (0.004) \\
					merge &	0.187 (0.001) &	0.280 (0.002) &	0.419 (0.002) &	\bf 0.614 (0.002) \\
					par-inject &	0.183 (0.004) &	0.275 (0.003) &	0.410 (0.003) &	0.605 (0.004) \\
					\hline\hline
				\end{tabular}
			\end{small}
		\end{minipage}
	}
	
	\subfloat[
		\label{tbl:results_gen_mscoco_2}Results for MSCOCO.
	]{
		\begin{minipage}{\textwidth}
			\centering
			\begin{small}
				\begin{tabular}{lcccc}
					\hline\hline
					&	BLEU-4 &	BLEU-3 &	BLEU-2 &	BLEU-1 \\
					init-inject &	\bf 0.271 (0.002) &	\bf 0.367 (0.002) &	\bf 0.502 (0.002) &	\bf 0.679 (0.003) \\
					pre-inject &	0.267 (0.002) &	0.366 (0.003) &	0.501 (0.003) &	0.677 (0.002) \\
					par-inject &	0.265 (0.003) &	0.359 (0.004) &	0.492 (0.004) &	0.667 (0.003) \\
					merge &	0.262 (0.003) &	0.362 (0.003) &	0.500 (0.003) &	0.677 (0.003) \\
					\hline\hline
				\end{tabular}
			\end{small}
		\end{minipage}
	}
\end{table}

Does merge's small size impact its performance when generated captions are compared to corpora? The metrics reported here show considerable variability in ranking of the various architectures depending on dataset. For example, CIDEr scores place init-inject at the top for both Flickr8K and MSCOCO, but merge outperforms it on this measure on Flickr30K. Comparing ROUGE-L, METEOR and CIDEr, init-inject seems to be ranked highest over most datasets (the situation is far more variable with the BLEU scores in Table~\ref{tbl:results_gen2}, however). However, the differences among architectures are very small. This is especially true for the larger MSCOCO dataset. Thus, though init-inject often comes out on top, the other architectures are not lagging behind by a wide margin. 

\subsection{Image retrieval}

Image retrieval results across the three datasets are shown in Table~\ref{tbl:results_ret}.

\begin{table}[t]
	\centering
	\caption{
		\label{tbl:results_ret}Results of the image retrieval metrics. Language model perplexity was also included here.
	}
	\subfloat[
		\label{tbl:results_ret_flickr8k}Results for Flickr8K.
	]{
		\begin{minipage}{\textwidth}
			\centering
			\begin{small}
				\begin{tabular}{lccccc}
					\hline\hline
					&	R@1 \% &	R@5 \% &	R@10 \% &	Med. rank &	Pplx. \\
					init-inject &	\bf 17.5 (0.6) &	\bf 43.2 (1.1) &	\bf 56.1 (0.4) &	\bf 7.7 (0.5) &	13.70 (0.02) \\
					merge &	17.3 (0.1) &	41.1 (0.4) &	54.6 (0.6) &	8.5 (0.4) &	13.96 (0.07) \\
					par-inject &	15.8 (0.4) &	39.7 (0.6) &	53.3 (0.6) &	9.0 (0.0) &	\bf 13.42 (0.08) \\
					pre-inject &	14.8 (0.2) &	37.9 (0.8) &	51.8 (1.1) &	9.3 (0.5) &	13.54 (0.13) \\
					\hline\hline
				\end{tabular}
			\end{small}
		\end{minipage}
	}
	
	\subfloat[
		\label{tbl:results_ret_flickr30k}Results for Flickr30K.
	]{
		\begin{minipage}{\textwidth}
			\centering
			\begin{small}
				\begin{tabular}{lccccc}
					\hline\hline
					&	R@1 \% &	R@5 \% &	R@10 \% &	Med. rank &	Pplx. \\
					merge &	\bf 24.2 (0.0) &	\bf 51.0 (0.5) &	\bf 61.5 (0.7) &	\bf 5.0 (0.0) &	22.08 (0.03) \\
					par-inject &	22.8 (0.8) &	48.1 (0.4) &	60.0 (0.9) &	6.0 (0.0) &	21.09 (0.46) \\
					init-inject &	22.7 (0.7) &	48.6 (0.3) &	60.1 (0.1) &	6.0 (0.0) &	\bf 19.50 (0.12) \\
					pre-inject &	21.5 (0.4) &	48.1 (0.7) &	60.0 (0.5) &	6.0 (0.0) &	20.18 (0.03) \\
					\hline\hline
				\end{tabular}
			\end{small}
		\end{minipage}
	}
	
	\subfloat[
		\label{tbl:results_ret_mscoco}Results for MSCOCO.
	]{
		\begin{minipage}{\textwidth}
			\centering
			\begin{small}
				\begin{tabular}{lccccc}
					\hline\hline
					&	R@1 \% &	R@5 \% &	R@10 \% &	Med. rank &	Pplx. \\
					init-inject &	\bf 29.1 (0.9) &	\bf 63.8 (0.9) &	\bf 77.2 (0.8) &	\bf 3.0 (0.0) &	\bf 9.27 (0.03) \\
					merge &	28.7 (0.7) &	62.2 (0.5) &	74.8 (0.5) &	\bf 3.0 (0.0) &	10.40 (0.04) \\
					par-inject &	27.2 (0.4) &	58.4 (0.7) &	73.3 (1.0) &	4.0 (0.0) &	10.07 (0.05) \\
					pre-inject &	27.1 (0.9) &	60.5 (0.6) &	75.1 (0.2) &	4.0 (0.0) &	9.88 (0.04) \\
					\hline\hline
				\end{tabular}
			\end{small}
		\end{minipage}
	}
\end{table}

When it comes to retrieving the most relevant image for a caption, we once again see merge ranked first on Flickr30K, while init-inject is at the top on Flickr8K and MSCOCO, on practically all $R$@$n$ measures, as well as median rank. Interestingly, in the two sets of cases where init-inject outperforms other architectures, merge is a close second, at least for $R$@$1$. In terms of perplexity, the general picture is in favour of inject models, with merge evincing marginally greater perplexity on all datasets. Overall, however, the outcomes mirror those of the previous sub-section: differences among architectures do not seem compelling and although the init-inject model outperforms merge in a number of instances, merge is a close second.

\subsection{Caption diversity metrics}

Next, we turn to the caption diversity metrics, shown in Table~\ref{tbl:results_div}.

\begin{table}[t]
	\centering
	\caption{
		\label{tbl:results_div}Results of the caption diversity metrics. The metrics were also applied to the first caption of each image in the human written test set captions.
	}
	\subfloat[
		\label{tbl:results_div_flickr8k}Results for Flickr8K.
	]{
		\begin{minipage}{\textwidth}
			\centering
			\begin{small}
				\begin{tabular}{lcccc}
					\hline\hline
					&	Vocab. Used \% &	Unigram Ent. &	Bigram Ent. &	Existing Caps. \% \\
					merge &	\bf 13.52 (0.97) &	5.591 (0.091) &	7.626 (0.159) &	11.03 (0.54) \\
					pre-inject &	12.46 (0.74) &	\bf 5.623 (0.064) &	\bf 7.672 (0.066) &	\bf 9.43 (0.31) \\
					init-inject &	12.01 (0.34) &	5.617 (0.043) &	7.649 (0.050) &	11.13 (0.05) \\
					par-inject &	10.67 (0.53) &	5.507 (0.023) &	7.421 (0.024) &	13.30 (2.20) \\
					\hline
					human &	46.75 (0.00) &	7.333 (0.000) &	10.836 (0.000) &	1.10 (0.00) \\
					\hline\hline
				\end{tabular}
			\end{small}
		\end{minipage}
	}
	
	\subfloat[
		\label{tbl:results_div_flickr30k}Results for Flickr30K.
	]{
		\begin{minipage}{\textwidth}
			\centering
			\begin{small}
				\begin{tabular}{lcccc}
					\hline\hline
					&	Vocab. Used \% &	Unigram Ent. &	Bigram Ent. &	Existing Caps. \% \\
					merge &	\bf 5.95 (0.13) &	5.368 (0.058) &	7.215 (0.099) &	\bf 6.77 (0.45) \\
					init-inject &	5.70 (0.11) &	5.509 (0.067) &	7.407 (0.078) &	8.70 (0.73) \\
					pre-inject &	5.45 (0.36) &	\bf 5.511 (0.045) &	\bf 7.438 (0.118) &	10.07 (2.34) \\
					par-inject &	3.77 (0.19) &	5.125 (0.028) &	6.720 (0.076) &	11.27 (1.10) \\
					\hline
					human &	29.40 (0.00) &	8.011 (0.000) &	11.786 (0.000) &	0.00 (0.00) \\
					\hline\hline
				\end{tabular}
			\end{small}
		\end{minipage}
	}
	
	\subfloat[
		\label{tbl:results_div_mscoco}Results for MSCOCO.
	]{
		\begin{minipage}{\textwidth}
			\centering
			\begin{small}
				\begin{tabular}{lcccc}
					\hline\hline
					&	Vocab. Used \% &	Unigram Ent. &	Bigram Ent. &	Existing Caps. \% \\
					merge &	\bf 7.91 (0.10) &	6.073 (0.023) &	8.738 (0.025) &	\bf 40.93 (0.11) \\
					init-inject &	7.26 (0.05) &	\bf 6.128 (0.009) &	\bf 8.768 (0.025) &	51.88 (0.20) \\
					pre-inject &	6.59 (0.16) &	6.064 (0.048) &	8.657 (0.049) &	51.92 (0.81) \\
					par-inject &	5.00 (0.05) &	5.863 (0.008) &	8.192 (0.028) &	62.73 (0.97) \\
					\hline
					human &	34.92 (0.00) &	7.833 (0.000) &	11.915 (0.000) &	7.14 (0.00) \\
					\hline\hline
				\end{tabular}
			\end{small}
		\end{minipage}
	}
\end{table}

These diversity metrics evince the most dramatic performance differences. If we focus on the proportion of generated captions that were found in the training set, on MSCOCO, this figure ranges from just over 40\% for merge to over 60\% for par-inject. With the exception of Flickr8K, merge has the lowest proportion of caption reuse overall. If these results are compared to those in preceding sub-sections, the fact that those models with the greatest tendency to reuse captions tend to perform well on corpus-based metrics such as CIDEr suggests that the datasets under consideration are highly stereotyped, perhaps with a significant amount of redundancy and lack of variety.

A similar observation has been made by \citet{Devlin2015}. In a comparison of retrieval-based and neural architectures for image captioning, these authors found that corpus-based metrics (especially BLEU) tend to give higher scores on test instances where the images were very similar to training instances. Neural architectures performed better for more similar images overall. 

The results obtained for the human captions (bottom rows of Table~\ref{tbl:results_div}) suggest that the level of caption reuse by humans is extremely low compared to the models under consideration, though it stands at 7\% on MSCOCO.

Turning to the extent to which architectures use their training vocabulary, the picture that emerges is consistent with the above. While humans used between 29\% and 47\% of the known vocabulary (taken from the training set) to describe the test set images, none of the evaluated systems used more than 14\%. The merge architecture tops the ranks for all datasets by a small margin, although unigram and bigram entropy is highest for pre-inject (Flickr8K and Flickr30K) and init-inject (MSCOCO). 

We interpret these results as showing that neural caption generators require seeing a word in the training set very often in order to learn to use it. From a methodological perspective, this further implies that setting an even higher frequency threshold, below which words are mapped to the UNKNOWN token (the current experiments set the threshold at five), would be feasible and would make relatively little difference to the results.

\subsection{Visual information retention}

As noted in Section~\ref{sec:introduction}, one of the differences between the architectures under consideration is whether they incorporate the image features early or late. This raises the possibility of differences in the degree to which visual information is retained by each architecture in the multimodal vector, that is, the input to `$\text{FF}^\text{out}$' in Figure~\ref{fig:implementation}. This is where information about visual and linguistic input is combined and is the information bottleneck that the output depends on. The question we want to answer is: Do (early-binding) inject architectures tend to `forget' about the image as more words are input into the RNN? Given that the RNN's memory is finite, it should be difficult to retain information about all inputs as the length of the sequence increases, so information about the image might start fading away as the input sequence gets longer. Merge architectures do not have this problem with visual information as it is kept outside of the RNN and so is fully retained in the multimodal vector regardless of the number of time steps.

To measure how much visual information is retained as the number of time steps grows, we do the following:
\begin{enumerate}
	\item Take a trained neural network and input an image and a matching caption.
	\item Record the multimodal vector in the neural network at every time-step.
	\item Replace the image from the original neural network in step 1 with a randomly selected image, paired with the original caption, thus introducing an image-caption mismatch.
	\item Record the new, adulterated multimodal vector at every time-step for the new caption-image combination.
	\item Compare the original and adulterated vectors: if these converge as more words are fed to the model, it implies that the multimodal vector is losing image information, as it would be getting influenced less by the image and more by the prefix.
\end{enumerate}

As a measure of distance between original and adulterated vectors, we use the mean absolute difference, that is, we take the absolute difference between each corresponding dimension in the two multimodal vectors and then take the mean of these differences. Mean absolute difference avoids giving a larger distance to larger vectors and is also intuitive as a measure of difference between vectors. It also keeps the distance between time steps exactly equal for merge, which is desirable since merge does not lose visual information across time steps.

\begin{figure}[t]
	\centering
	\includegraphics[scale=0.75]{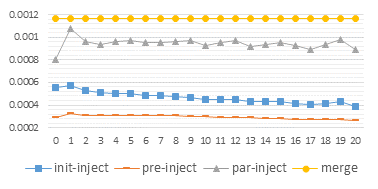}
	\caption{
		\label{fig:multimodvec_image_memory}
		Difference between multimodal vectors over time, when the caption is held constant but the image is changed. For a trained model, markers show mean absolute difference between multimodal vectors when a 20-word caption is fed to the model with the original, versus a randomly chosen image. The marker at position 0 gives the multimodal vector difference after feeding in the START token whilst the marker at position 20 gives the multimodal vector difference after feeding in the last word in the caption.
	}
\end{figure}

For this set of experiments, we used all 20-word captions in the MSCOCO test set and measured the mean distance over all 21 time steps (the 20 words plus the START token). 20-word captions are long enough to see a trend without ending up with too few captions (the mean caption length on the MSCOCO test set is about $10.4$). To create a more reliable mean we repeat this procedure 100 times so that the mean is over all images in the test set using 100 random images per instance. The results are shown in Figure~\ref{fig:multimodvec_image_memory}.

None of the inject architectures maintained a consistent distance between the original and adulterated multimodal vectors. Crucially, the merge architecture also has the largest distance among all architectures, demonstrating that, in this architecture, the words in a caption exhibit a greater dependency on the image to which they pertain (hence, adulterating the multimodal vector with an irrelevant image alters the representation considerably). Par-inject comes in second place in terms of multimodal vector distance. This suggests that it retains more visual information than the other inject architectures, though not as much as merge. It seems that the amount of retention across time steps changes somewhat unpredictably, but tends to decrease overall, which means that information gets lost over time (though not to the extent of init-inject and pre-inject). Init-inject comes third in visual information retention followed by pre-inject, both of which decrease over time. It seems that, in a GRU trained for caption generation, the initial hidden state vector exerts more influence on the final hidden state vector than the first input.

These results predict that if the generated captions needed to be very long, late binding architectures will produce better captions as they will retain visual information over longer time steps, maintaining a tighter coupling between visual and linguistic information.

%% file: tex/conclusion.tex
\section{Conclusion}
\label{sec:conclusion}

This paper presented a systematic evaluation of a number of variations on architectures for image caption generation and retrieval. The primary focus was on the distinction between what we have termed `inject' and `merge' architectures. The former type of model mixes image and language information by training an RNN to encode an image-prefix mixture. By contrast, merge architectures maintain a separation between an RNN subnetwork, which encodes a linguistic string, and the image vector, merging them late in the process, prior to a prediction step. These models are therefore compatible with approaches to image caption generation using a `multimodal' layer \citep{Mao2014,Mao2015,Mao2015a,Hendricks2016}. While both types of architectures have been discussed in the literature, the inject architecture has been more popular. 

Yet, there has been little systematic evaluation of its advantages compared to merge. Our experiments show that on standard corpus-based metrics such as CIDEr, the difference in performance between architectures is rather small. Init-inject tends to be better at generation and retrieval measures. Thus, from the perspective of corpus similarity, early binding of image features in models that view such features as ``modifiable'' (in the sense outlined in the introduction) appear to be better than the alternatives. 

Crucially, however, we also show that inject architectures are much more likely to re-generate captions wholesale from the training data and evince less vocabulary variation. Hence, from the perspective of variation, late-binding models that treat image features as fixed (i.e. not mixed with linguistic features) are better. While this is due in part to the nature of the available corpora, the superior performance of merge on this measure does suggest that, by encoding information from the two modalities separately, merge architectures might be producing less generic and stereotyped captions, exploiting their multimodal resources more effectively.

Our experiments on visual information retention show that, over time, inject architectures tend to loosen the coupling between visual and linguistic features, so that the difference between actual and adulterated multimodal vectors gets smaller. This too supports the view that inject models may, especially for longer captions, tend towards more generic and less image-specific captions, a finding that echoes the observations of \citet{Devlin2015}, to some extent. In any case, late merging is, by definition, not susceptible to this problem.

From an engineering perspective, there is a significant difference between the required sizes of the RNN hidden state vectors. Whilst merge only requires a hidden state vector size sufficient to `remember' caption prefixes, which depends on the length and complexity of the training set captions, inject architectures require additional memory to also store image information. This means that merge architectures make better use of their RNN memory. They also require less regularization whilst maintaining similar performance as other architectures.

The work presented here opens up some avenues for future research. In future work, we hope to investigate whether the results in this paper would remain similar when the experiments are repeated on other applications of conditioned neural language models such as neural machine translation or question answering.

Furthermore, by keeping language and image information separate, merge architectures lend themselves to potentially greater portability and ease of training. For example, it should be possible in principle to take the parameters of the RNN and embedding layers of a general text language model and transfer them to the corresponding layers in a caption generator. This would reduce training time as it would avoid learning the RNN weights and the embedding weights of the caption generator from scratch. As understanding of deep learning architectures evolves in the NLP community, one of our goals should be to maximise the degree of transferability among model components.